\documentclass[12pt]{article}
\usepackage{amsmath}
\usepackage{amssymb}
\usepackage{graphicx,psfrag,epsf}
\usepackage[mathscr]{euscript}
\usepackage{enumerate}
\usepackage{url} 
\usepackage{natbib}
\usepackage{amsmath, amsthm, amssymb}

\usepackage{amssymb}
\usepackage{bbm}
\usepackage[table]{xcolor} 
\usepackage{algorithm}
\usepackage[noend]{algpseudocode}
\usepackage{comment}
\usepackage{multirow}
\usepackage{comment}
\usepackage[skip=0.5\baselineskip]{caption}
\usepackage{float}
\floatstyle{plaintop}
\restylefloat{table}

\begin{document}

\def\spacingset#1{\renewcommand{\baselinestretch}%
{#1}\small\normalsize} \spacingset{1}


\title{\bf Tensor-on-tensor Regression Neural Networks for Process Modeling with High-dimensional Data}
\author{Qian Wang, Mohammad N. Bisheh and Kamran Paynabar\\
\\
    School of Industrial and Systems Engineering, \\
    Georgia Institute of Technology, Atlanta, Georgia}
\maketitle
\begin{abstract}
  Modern sensing and metrology systems now stream terabytes of heterogeneous, high-dimensional (HD) data profiles, images, and dense point clouds, whose natural representation is multi-way tensors. Understanding such data requires regression models that preserve tensor geometry, yet remain expressive enough to capture the pronounced nonlinear interactions that dominate many industrial and mechanical processes. Existing tensor-based regressors meet the first requirement but remain essentially linear. Conversely, conventional neural networks offer nonlinearity only after flattening, thereby discarding spatial structure and incurring prohibitive parameter counts. This paper introduces a Tensor-on-Tensor Regression Neural Network (TRNN) that unifies these two paradigms. TRNN adopts an autoencoder inspired encoder–decoder architecture whose learnable shrinking and expanding Tucker layers preserve multilinear structure throughout the network, while a novel contraction operator at the bottleneck enables mappings between tensors of different orders and injects layer-wise nonlinearity. We derive closed form forward and back-propagation formulae, show that a single layer linear special case reduces to partial least squares, and provide guidance on rank selection for practical training. Extensive simulations demonstrate that TRNN reduces relative mean square error by up to 45\% compared with state-of-the-art (SOTA) linear tensor regressors and achieves comparable or faster runtimes than flattened neural baselines. Two real world case studies of geometric error prediction in the point cloud modeling in a truning process as well as a curve-on-curve modeling in an vehicle engine emission control systems were performed to illustrate TRNN’s ability to deliver accurate, interpretable predictions from HD process data while scaling efficiently to industrial resolutions.

\end{abstract}

\noindent%
\vfill

\newpage
\spacingset{1.45} 

\section{Introduction}
Modern production environments, especially those embracing additive and smart manufacturing paradigms, now operate under an unprecedented severe flood of measurement data. Dense arrays of vibration and acoustic sensors stream functional signals in real time \cite{lei2010automatic, yan2018real}, in situ vision systems capture micron resolution images at every critical stage \cite{gibson2021design}, and coordinate-measurement machines or structured light scanners routinely generate full 3-D point clouds with submillimeter precision \cite{jin2000identification, prasath2013comparison, yan2017anomaly}. These heterogeneous high-frequency measurements yield high-dimensional (HD) functional data that far exceed the capacity of classical multivariate models \cite{yan2019structured}. Crucially, each modality samples a distinct facet of the underlying physics, so integrating them promises deeper process understanding, earlier fault detection, and tighter quality control than treating any single stream in isolation.

\begin{figure}[htbp]
\includegraphics[scale=0.5]{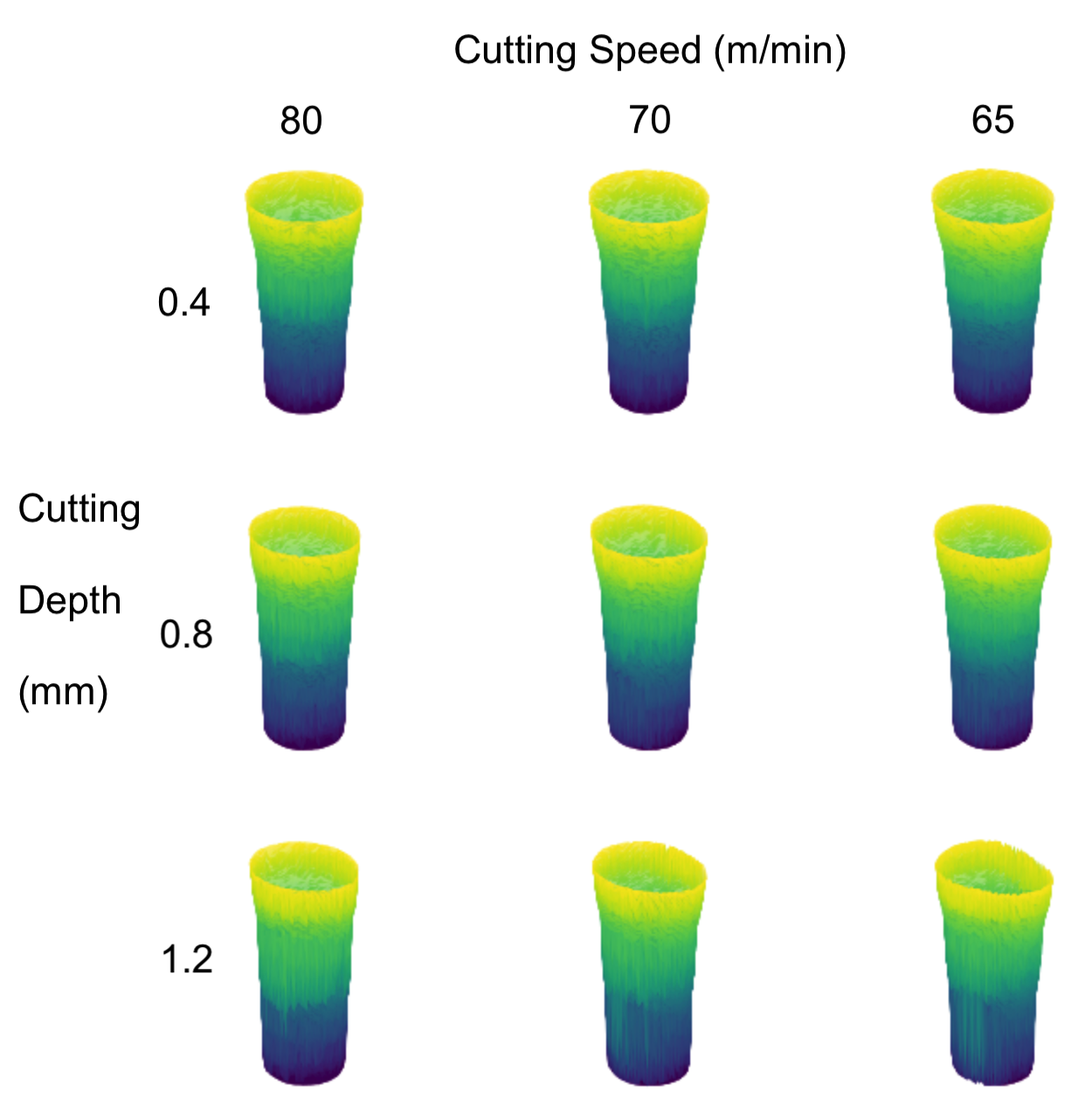}
\centering
\caption{\label{fig:case1plot} Examples of average cylinders under different cutting conditions}
\end{figure}

Realizing that promise demands statistical tools capable of learning mappings between heterogeneous HD predictors and responses while preserving the multi-way structure of every data stream. Such requirements arise across a broad spectrum of applications. Semiconductor manufacturing, for example, predicts wafer overlay errors by regressing future shape point clouds on lithography images \cite{turner2013role}. In precision turning, the machining parameters influence the geometric accuracy of the finished parts, necessitating tensor-to-surface regression models \cite{pacella2018multilinear}. Figure \ref{fig:case1plot} visualizes this challenge: the average point clouds of cylindrical workpieces produced under three representative cutting conditions reveal systematic, spatially complex deviations from a nominal cylinder. Successful process models must therefore capture nonlinear, cross-modal dependencies between the process setting tensor (speed, feed, depth of cut, vibration signatures) and the response tensor (dense 3-D geometry).

Analogous HD–HD relationships permeate materials science, where microstructure images are linked to thermomechanical histories to optimize alloy performance \cite{khosravani2017development, gorgannejad2019quantitative}; in food engineering, where hyperspectral cubes predict moisture and texture distributions \cite{yu2003multivariate}; and in structural health monitoring, where guided wave images are regressed on environmental variables to track damage progression \cite{bellon1995experimental, balageas2010structural}. Across these domains, two technical hurdles recur: (i) the covariates and responses are structured tensors whose inherent geometry must be respected, and (ii) their relationships are often highly non-linear, rendering traditional linear-factor approaches inadequate. Addressing these challenges and motivated by the illustrative machining example in Figure \ref{fig:case1plot}, this work develops a modeling framework that jointly leverages tensor structure and deep hierarchical nonlinearity to deliver accurate and computationally tractable HD regression.

\section{Related work}

Extensive research has focused on regression analysis for scalar or one-dimensional functional data, especially within the framework of functional data analysis (FDA). Classic approaches include functional principal component regression (FPCR) and partial least squares (PLS), which reduce dimensionality by projecting high-dimensional profiles into subspaces where traditional regression can be effectively applied  \cite{liang2003relationship, ramsay2008functional, yao2005functional, reiss2010fast, fan2014functional, ivanescu2015penalized, luo2017function}. These methods mitigate overfitting due to high dimensionality but often assume linearity and require prespecified basis functions, limiting their ability to model complex nonlinear relationships.

To overcome this limitation, recent FDA research has introduced data-driven basis learning techniques, such as penalized or adaptive splines \cite{fan2014functional, luo2017function}, yet challenges persist when generalizing to high-dimensional, unstructured data like images or 3D point clouds \cite{colosimo2011analyzing}. A notable early solution was projecting structured profiles into principal component spaces prior to statistical analysis (e.g., multivariate ANOVA), but such approaches ignore interactions between process variables and spatial structure.

Tensor analysis provides a powerful alternative for modeling structured high-dimensional data. Tensor representations preserve spatial correlations across multiple dimensions and have been adopted in domains such as quality control \cite{sun2006window}, surveillance \cite{sapienza2015detecting}, and medical imaging \cite{yan2014image}. Tensor regression models, particularly those using Canonical Polyadic (CP) and Tucker decompositions, have shown promise in scalar-on-tensor applications \cite{zhou2013tensor, li2018tucker}. These models approximate the parameter space via low-rank decompositions to manage the curse of dimensionality.

However, much of this work has focused on scalar response models. Early attempts to generalize tensor regression to tensor responses (tensor-on-tensor regression) include multiscale spatial modeling \cite{li2011multiscale}, parsimonious modeling of structured outputs \cite{li2017parsimonious}, and separable basis expansions \cite{yan2019structured}. Despite their utility, these methods often impose structural assumptions (e.g., separability or matching rank across input-output modes) or suffer from limited expressiveness.

True tensor-on-tensor regression, where both inputs and outputs are tensors of potentially different ranks, introduces unique challenges. The foundational work of \cite{bro1996multiway} on multi-way PLS (N-PLS) laid the groundwork for multilinear extensions of PLS, later formalized in higher-order PLS (HOPLS) \cite{zhao2012higher}. These methods enable regression across coupled tensor modes, but are typically linear and computationally expensive for large-scale applications.

More recently, Lock (2018) and Gahrooei et al. (2021) introduced frameworks for multilinear tensor-on-tensor regression, leveraging tensor decomposition to reduce complexity while preserving interpretability \cite{lock2018tensor, gahrooei2021multiple}. However, these models still rely on linear transformations and are thus limited in their ability to model nonlinear relationships common in real-world systems.

Neural networks offer a flexible alternative, particularly suited for capturing nonlinear interactions where traditional statistical models fail. Extensions to tensor input have included Tensor Neural Networks (TNNs) \cite{liu2016tensor}, 3D CNNs \cite{li20213d}, and deep tensor autoencoders (DTAEs) \cite{qian2021dtae}. \cite{qian2021dtae} proposed a tensorized version of backpropagation (TBP) to enable deep autoencoders training on seismic data, showing superiority over classical low-rank tensor completion when the low-rank assumption is violated.

Other architectures, such as the Tensor Regression Networks (TRNs) by \cite{kossaifi2017tensor}, introduced parameter efficient deep nets with embedded low-rank priors for scalar outputs. However, most of these neural approaches focus on classification or scalar prediction and are not tailored for full tensor-on-tensor mappings. Flattening high-dimensional inputs into vectors often leads to parameter explosion and loss of spatial structure \cite{chien2017tensor}.

Recent progress in tensor‐based learning is exemplified by the factor augmented tensor‐on‐tensor neural network of \citet{zhou2025factor}, which first compresses each covariate tensor through a linear Tucker factor model and subsequently predicts future tensors with a temporal convolutional network.  While their model achieves notable speed and accuracy on large spatio-temporal streams, its linear front-end inevitably removes non-linear interactions that occur within individual snapshots, forces the input and output tensors to share the same order, and restricts the model’s expressive power to the temporal stage handled by the downstream network.

Consequently, there remains a methodological gap for approaches that can perform fully non-linear tensor-on-tensor regression, preserve multi-way structure at every layer, and remain computationally tractable for the extremely high dimensionalities encountered in industrial settings.  To close this gap, we develop a Tensor-on-Tensor Regression Neural Network (TRNN) whose architecture adopts a compact encoder–decoder architecture inspired by autoencoders, employs learnable shrinking and expanding Tucker layers that respect multilinear geometry end-to-end, and introduces a novel contraction operator at the bottleneck that permits mappings between tensors of different orders while injecting additional non-linearity through Rectified Linear Unit (ReLU) activations.  To the best of our knowledge, TRNN is the first non-linear neural framework that addresses heterogeneous tensor-on-tensor regression problems in the industrial-engineering domain.

The remainder of the paper is organized as follows.  Section~\ref{sec:Methodology} presents the TRNN architecture in detail, derives the corresponding forward and backward propagation formulae, and shows that a single-layer linear special case reduces to PLS.  Section~\ref{sec:Simulation} uses two simulation studies to benchmark TRNN against SOTA linear and deep baselines, and Section~\ref{sec:Casestudy} reports two real-world case studies, one on turning process geometry prediction and the other on microstructure process mapping to demonstrate the method’s practical benefits. Finally, Section~\ref{sec:Conclusion} summarizes the contributions and outlines the directions for future research.\\

\section{Methodology}
\label{sec:Methodology}
In this section, we proposed our TRNN methodology to model the processes which involve a HD input tensor consisting of multiple signals and a HD output tensor representing the observed structured functional values given the signals. Given this model, the prediction of the functional output given the control signals becomes possible. 

\subsection{Proposed Network Structure and Overview of Tensor Concepts and Operators}
\label{sec:Overview}

A tensor is a multi-dimensional array; its order (also called mode or way) is the number of dimensions it possesses.  A vector is thus a first-order tensor and a matrix a second-order tensor.  For a general $N$-th-order tensor $\mathcal{X}\in\mathbb{R}^{I_{1}\times I_{2}\times\cdots\times I_{N}}$, the Frobenius norm 
$$
\|\mathcal{X}\|_F=\sqrt{\sum_{i_{1} i_{2} \cdots i_{N}} \mathcal{X}_{i_{1} i_{2} \cdots i_{N}}^{2}}
$$
\vspace{0.7em}
\noindent
is $n$-mode product.  
Let $\mathbf{U}\in\mathbb{R}^{J\times I_{n}}$ be a matrix and
$\mathcal{X}\in\mathbb{R}^{I_{1}\times\cdots\times I_{n}\times\cdots\times I_{N}}$ a tensor.  
Their $n$-mode product, denoted $\mathcal{X}\times_{n}\mathbf{U}$, is defined by
\[
(\mathcal{X}\times_{n}\mathbf{U})_{i_{1}\cdots i_{n-1}\,j\,i_{n+1}\cdots i_{N}}
      =\sum_{i_{n}=1}^{I_{n}}
        \mathcal{X}_{i_{1}\cdots i_{n}\cdots i_{N}}
        \,\mathbf{U}_{j\,i_{n}},
\]
and yields a tensor in $\mathbb{R}^{I_{1}\times\cdots\times I_{n-1}\times J\times I_{n+1}\times\cdots\times I_{N}}$.  

\vspace{0.7em}
\noindent
Let $\mathcal{X}\in\mathbb{R}^{P_{1}\times\cdots\times P_{\ell}}$ and
$\mathcal{C}\in\mathbb{R}^{P_{1}\times\cdots\times P_{\ell}\times Q_{1}\times\cdots\times Q_{d}}$.  
Their contraction,
$\mathcal{X}*\mathcal{C}\in\mathbb{R}^{Q_{1}\times\cdots\times Q_{d}}$, is
\[
(\mathcal{X}*\mathcal{C})_{q_{1}\cdots q_{d}}
      =\sum_{p_{1}=1}^{P_{1}}\!\cdots\!\sum_{p_{\ell}=1}^{P_{\ell}}
        \mathcal{X}_{p_{1}\cdots p_{\ell}}
        \,\mathcal{C}_{p_{1}\cdots p_{\ell}\,q_{1}\cdots q_{d}}.
\]

\vspace{0.7em}
\noindent
Tucker decomposition expresses a tensor as a multilinear product of a (usually much smaller) core tensor and mode-specific factor matrices:
\[
\mathcal{X}\;\approx\;
\mathcal{G}\times_{1}\mathbf{U}_{1}\times_{2}\mathbf{U}_{2}\cdots\times_{N}\mathbf{U}_{N},
\qquad
\mathcal{G}\in\mathbb{R}^{r_{1}\times r_{2}\times\cdots\times r_{N}},\;
\mathbf{U}_{k}\in\mathbb{R}^{I_{k}\times r_{k}},
\]
with typically $r_{k}\ll I_{k}$.  
This decomposition dramatically reduces parameter count  and is reflected in the encoder–decoder design that follows.

\vspace{0.7em}

\begin{figure}[htbp]
\includegraphics[scale=0.5]{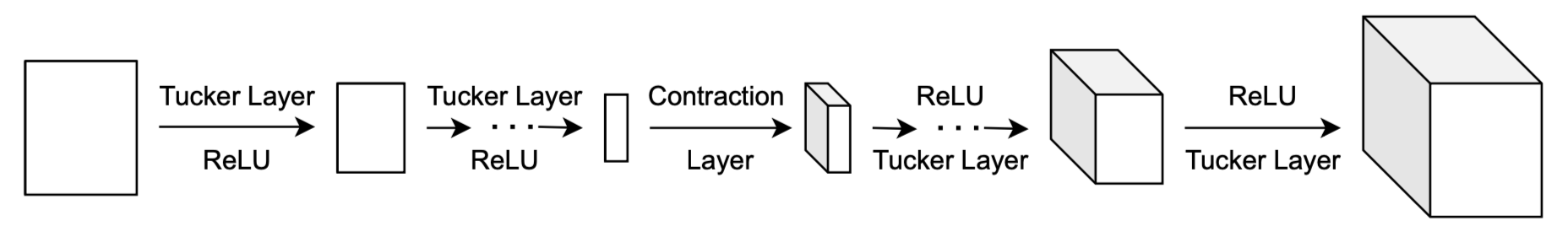}
\centering
\caption{\label{fig:nnstru}The network struture of the proposed TRNN}
\end{figure}

Assume the size of data is $N$, the $l$-order input tensor $X\in \mathbb{R}^{N\times P_2\times P_3\times...\times P_l}$ and the $d$-order output tensor $Y\in \mathbb{R}^{N\times Q_2\times Q_3\times...\times Q_d}$. The aim of the TRNN framework is to build a non-linear regression model between $X$ and $Y$ using the expressive power of neural networks to capture their complex relationship and reduce the prediction errors. The structure of the proposed neural networks is demonstrated in Figure \ref{fig:nnstru}.\\

As illustrated in Figure~\ref{fig:nnstru}, the proposed network architecture consist of two main components connected by a central contraction layer. The left side of the network serves as an encoder that progressively reduces the dimensionality of the input tensor through Tucker layers equipped with ReLU activations. This compression culminates in the contraction layer, which not only transforms the latent representation but may also change the tensor order. The right side of the network functions as a decoder, symmetrically expanding the compressed representation to match the dimensions of the output tensor. This design mirrors the well-established autoencoder architecture~\citep{https://doi.org/10.1002/aic.690370209}, where information is minimally compressed at the bottleneck to retain only the most essential features before reconstruction. Two key distinctions set our approach apart: (i) we apply mode-$n$ tensor products to reduce or expand dimensions rather than vector-based operations, and (ii) the contraction layer enables flexible mappings between input and output tensors with potentially different orders, a capability not typically supported in conventional autoencoders.\\

The motivation for adopting an autoencoder-like structure is twofold. First, performing contraction directly between the high-dimensional tensors $X$ and $Y$ is computationally expensive, often redundant, and prone to overfitting due to the large number of parameters involved. By reducing their dimensionality prior to contraction, the network becomes more parameter-efficient and less susceptible to overfitting. Second, high-dimensional data in industrial and physical systems often lie near a lower-dimensional manifold. By nonlinearly compressing both the input and output tensors through Tucker layers before contraction, the network exploits this intrinsic low-rank structure. As a result, a compact contraction layer at the bottleneck is sufficient to capture the complex nonlinear relationships between $X$ and $Y$.\\

To frame the above panorama, we define the forward and backward propagation rules associated with each layer in the proposed network. These derivations enable the application of standard neural network training procedures, including mini-batch stochastic gradient descent (SGD)~\citep{saad1998online} and more advanced optimizers such as Adam~\citep{kingma2014adam}. Once the propagation rules are established, the network parameters can be updated end-to-end using backpropagation in a fully differentiable framework.\\

The proposed architecture consists of an encoder that progressively compresses each mode of the input tensor $\mathcal{X}$ using Tucker layers combined with ReLU activations, followed by a contraction layer that both alters the tensor order and serves as an information bottleneck. This is succeeded by a decoder that symmetrically reconstructs the compressed representation to match the dimensional structure of the output tensor $\mathcal{Y}$. By reducing dimensionality prior to contraction, the network effectively mitigates overfitting and leverages the intrinsic low-rank structure frequently observed in HD process data. The following Sections formulate the forward and backward propagation rules associated with each operator, enabling end-to-end training via standard optimization algorithms such as mini-batch stochastic gradient descent~\citep{saad1998online} or Adam~\citep{kingma2014adam}.\\

\subsection{Forward Propagation}
\label{sec:ForwardProp}

We denote the input tensor as $s_0 = \mathcal{X}$. The forward propagation process consists of a sequence of Tucker-based transformations and nonlinear activations, organized symmetrically around a central contraction layer, and concluding with the prediction $\hat{Y}$.

The forward pass begins with a series of shrinking Tucker layers, which reduce the spatial dimensions of the input tensor by applying mode-wise tensor–matrix multiplications. Let the input to the $n$-th shrinking Tucker layer be 
\[
s_{n-1} \in \mathbb{R}^{N \times P_2^{(n-1)} \times P_3^{(n-1)} \times \cdots \times P_\ell^{(n-1)}},
\]
and the output be 
\[
r_n \in \mathbb{R}^{N \times P_2^{(n)} \times P_3^{(n)} \times \cdots \times P_\ell^{(n)}}, \quad \text{where } P_k^{(n)} < P_k^{(n-1)}.
\]
the formula for the $n^{th}$ shrinking Tucker layer is:
\begin{equation*}
    r_n = s_{n-1} \times_2 U_2^{(n)} \times_3 \cdots \times_\ell U_\ell^{(n)},
\end{equation*}
where $U_k^{(n)} \in \mathbb{R}^{P_k^{(n)} \times P_k^{(n-1)}}$ are learnable parameters.

Each Tucker layer is followed by a ReLU activation layer, which introduces nonlinearity through an element-wise rectified linear unit. Let the input of the $n^{th}$ shrinking ReLU layer is $r_{n}\in \mathbb{R}^{N\times P_2^{(n)}\times P_3^{(n)}\times...\times P_l^{(n)}}$ and output is $s_n\in \mathbb{R}^{N\times P_2^{(n)}\times P_3^{(n)}\times...\times P_l^{(n)}}$, the formula is: 
\begin{equation*}
    s_n = \max(r_n, 0).
\end{equation*}

After the sequence of Tucker and ReLU layers, the resulting tensor $s_{n_1}$ is passed to the contraction layer, which functions as both an information bottleneck and a mechanism to reduce or transform the tensor’s order. This layer produces an intermediate representation
\[
z_0 \in \mathbb{R}^{N \times Q_2^{(0)} \times Q_3^{(0)} \times \cdots \times Q_d^{(0)}}
\]
via the Einstein contraction product:
\begin{equation*}
    z_0 = s_{n_1} * \mathcal{C},
\end{equation*}
where $\mathcal{C} \in \mathbb{R}^{P_2^{(n_1)} \times \cdots \times P_\ell^{(n_1)} \times Q_2^{(0)} \times \cdots \times Q_d^{(0)}}$ is a learnable core tensor. The contraction operation effectively integrates the compressed features from all input modes and maps them into a latent representation aligned with the structure of the output tensor.\\

Following the contraction, a sequence of expanding ReLU layers and expanding Tucker layers reconstruct the output tensor structure. Let the input of the $n^{th}$ expanding ReLU layer is $z_{n}\in \mathbb{R}^{N\times Q_2^{(n)}\times Q_3^{(n)}\times...\times Q_d^{(n)}}$ and output is $a_n\in \mathbb{R}^{N\times Q_2^{(n)}\times Q_3^{(n)}\times...\times Q_d^{(n)}}$, $a_n$ is defined as:
\begin{equation*}
    a_n = \max(z_n, 0),
\end{equation*}
and the subsequent Tucker layer expands the dimensions as:
\begin{equation*}
    z_n = a_{n-1} \times_2 W_2^{(n)} \times_3 \cdots \times_d W_d^{(n)},
\end{equation*}
where $a_{n-1} \in \mathbb{R}^{N \times Q_2^{(n-1)} \times \cdots \times Q_d^{(n-1)}}$, 
$z_n \in \mathbb{R}^{N \times Q_2^{(n)} \times \cdots \times Q_d^{(n)}}$, and 
$W_k^{(n)} \in \mathbb{R}^{Q_k^{(n)} \times Q_k^{(n-1)}}$ are trainable parameters with $Q_k^{(n)} > Q_k^{(n-1)}$.\\

After $n_2$ expanding Tucker layers, the network produces its output $\hat{Y} = z_{n_2}$. The training loss is defined as the mean squared error between the predicted and true output tensors:
\begin{equation*}
    \mathcal{E}=\frac{1}{2 N} \sum_{i q_{2} \cdots q_{d}}(\hat{Y}-Y)^{2}_{iq_{2} \cdots q_{d}}
\end{equation*}

\subsection{Backward Propagation}
\label{sec:Backwardprop}

Based on the definitions of the $k$-mode product and the contraction product, the following gradient expressions can be derived for backpropagation through each layer of the proposed architecture. The derivative of the mean squared error with respect to the network output $z_{n_2}$ (Loss function) is given by:
\begin{equation*}
    \frac{\partial \mathcal{E}}{\partial z_{n_{2}i q_{2} \cdots q_d}} = \frac{1}{N} \left(z_{n_2} - Y\right)_{i q_2 \cdots q_d}.
\end{equation*} \\
The gradients for the expanding Tucker layer with ReLU are given by:
\begin{equation*}
    \frac{\partial \mathcal{E}}{\partial z_{n-1i\tilde{q_2}\cdots \tilde{q_{d}}}}=\sum_{q_2\cdots q_d} \frac{\partial \mathcal{E}}{\partial z_{ni q_{2} \ldots q_{d}} }W^{(n)}_{2\ q_2\tilde{q_2}}\cdots W^{(n)}_{d\ q_d\tilde{q_d}}\mathbbm{1}{\{z_{n-1i\tilde{q_2} \ldots \tilde{q_d}}\geqslant0\}}
\end{equation*}
\begin{equation*}
    \frac{\partial \mathcal{E}}{\partial W^{(n)}_{k\ q_{k}\tilde{q_{k}}}}=\sum_{i q_2\cdots q_{k-1}q_{k+1}\cdots q_d} \frac{\partial \mathcal{E}}{\partial z_{n i q_{2} \cdots q_{d}}} \sum_{ \tilde{q_{2}}\cdots \widetilde{q_{k-1}}\widetilde{q_{k+1}}\cdots \tilde{q_d}} a_{n-1 i \tilde{q_2} \cdots \tilde{q_{d}}} W^{(n)}_{2 q_2\tilde{q_2}} W^{(n)}_{k-1 q_{k-1}\widetilde{q_{k-1}}} W^{(n)}_{k+1 q_{k+1}\widetilde{q_{k+1}}} W^{(n)}_{d q_d\tilde{q_d}}
\end{equation*}
\\
Gradients with respect to the input $s_{n_1}$ and the contraction core $\mathcal{C}$ are:
\begin{equation*}
    \frac{\partial \mathcal{E}}{\partial s_{n_1,\,i\,p_2 \cdots p_\ell}} = 
    \sum_{q_2 \cdots q_d}
    \frac{\partial \mathcal{E}}{\partial z_{0,\,i\,q_2 \cdots q_d}} 
    \cdot \mathcal{C}_{p_2 \cdots p_\ell\,q_2 \cdots q_d},
\end{equation*}

\begin{equation*}
    \frac{\partial \mathcal{E}}{\partial \mathcal{C}_{p_2 \cdots p_\ell\,q_2 \cdots q_d}} = 
    \sum_i 
    \frac{\partial \mathcal{E}}{\partial z_{0,\,i\,q_2 \cdots q_d}} 
    \cdot s_{n_1,\,i\,p_2 \cdots p_\ell}.
\end{equation*}\\

For the shrinking layers on the encoder side, the gradients are:
\begin{equation*}
    \frac{\partial \mathcal{E}}{\partial r_{n-1 i \tilde{p_2} \cdots \tilde{p_l}}}=\sum_{p_{2} \cdots p_{l}} \frac{\partial \mathcal{E}}{\partial r_{n i p_{2} \ldots p_{l}}} U_{2\ p_{2}\tilde{p_2}}^{(n)} \cdots U_{l\ p_{l}\tilde{p_l}}^{(n)} \mathbbm{1}\left\{r_{n-1 i \tilde{p_{2}} \cdots \tilde{p_l}} \geqslant 0\right\}
\end{equation*}
\begin{equation*}
    \frac{\partial \mathcal{E}}{\partial U^{(n)}_{k\ p_{k}\tilde{p_{k}}}}=\sum_{i p_2\cdots p_{k-1}p_{k+1}\cdots p_l} \frac{\partial \mathcal{E}}{\partial r_{n i p_{2} \cdots p_{l}}} \sum_{ \tilde{p_{2}}\cdots \widetilde{p_{k-1}}\widetilde{p_{k+1}}\cdots \tilde{p_l}} s_{n-1 i \tilde{p_2} \cdots \tilde{p_{l}}} U^{(n)}_{2 p_2\tilde{p_2}} U^{(n)}_{k-1 p_{k-1}\widetilde{p_{k-1}}} U^{(n)}_{k+1 p_{k+1}\widetilde{p_{k+1}}} U^{(n)}_{l p_l\tilde{p_l}}
\end{equation*}
\\
Once these gradients have been computed, the network parameters 
$\{U_k^{(n)}, \mathcal{C}, W_k^{(n)}\}$ can be updated using standard backpropagation. Optimization can be performed using first-order methods such as mini-batch stochastic gradient descent~\citep{saad1998online} or adaptive algorithms such as Adam~\citep{kingma2014adam}.

\subsection{Connection with Partial Least Square}
\label{sec:ConnectionPLS}

A notable special case of the proposed TRNN architecture arises under the following conditions: (i) both the input and output tensors are of order two (i.e., matrices, where one mode indexes samples and the other features), (ii) all ReLU activation layers are removed, and (iii) the encoder and decoder each consist of a single linear layer. We refer to this simplified model as the Single-Layer Linear TRNN (SL-TRNN). Under this configuration, the training objective reduces to the following matrix regression problem:
\begin{equation*}
    \min\limits_{W,\,B,\,V} \;\; \|Y - XWBV^{\top}\|_F^2,
\end{equation*}
where $W$, $B$, and $V$ denote the parameter matrices of the left linear layer, the contraction layer, and the right linear layer, respectively.\\

PLS ~\citep{wold2001pls} is a classical linear regression method that models the relationship between an input matrix $X$ and a response matrix $Y$ by projecting both onto lower-dimensional latent subspaces. The goal of PLS is to find a projection of $X$ that best explains the variance in a corresponding projection of $Y$. Specifically, PLS identifies latent score matrices $T$ and $U$ along with corresponding loading matrices $W$ and $V$, such that the projections of $X$ and $Y$ maximize the covariance between $T$ and $U$. The decomposition is given by:
\begin{equation*}
\begin{aligned}
X &= T W^{\top} + E, \\
Y &= U V^{\top} + F,
\end{aligned}
\end{equation*}
where $E$ and $F$ denote the residual error matrices. Unlike ordinary least squares, which seeks to minimize prediction error only in $Y$, PLS simultaneously considers the variation in both $X$ and $Y$ through latent variable modeling, making it especially useful when the predictors are highly collinear or when the number of features exceeds the number of samples.\\

After the projection matrices $W$ and $V$ have been learned such that $T = XW$ and $U = YV$, the regression between the latent variables $T$ and $U$ is modeled via a coefficient matrix $B$. The optimization problem for learning $B$ becomes:
\begin{equation*}
    \min\limits_{B} \;\; \|YV - XWB\|_F^2,
\end{equation*}
which seeks to minimize the difference between the projected response and the linear transformation of the projected input.

If the projection matrix $V$ is orthogonal (i.e., $V^{\top}V = I$), this formulation is equivalent to:
\begin{equation*}
    \min\limits_{B} \;\; \|Y - XWBV^{\top}\|_F^2,
\end{equation*}
which matches the optimization form used in the SL-TRNN model described earlier. This connection highlights how SL-TRNN generalizes PLS by interpreting it as a special case within a neural tensor regression framework.\\

\section{Simulation Studies}
\label{sec:Simulation}

In this section, we did two simulation studies to evaluate the performance of our algorithm. The first study focused on scalar-on-tensor regression, and we compared our approach with the corresponding SOTA tensor-based method OTDR \citep{yan2019structured}. The second focused on tensor-on-tensor regression, and we compared our approach with the corresponding SOTA tensor-based method MTOT \citep{gahrooei2021multiple}. Also, to have fair comparison, we compare our performance with the benchmark from neural networks side, using the simplest vanilla neural networks (Vanilla NN).

\subsection{Water Drop Point Cloud Simulation}
\label{sec:Waterdrop}

The water drop point clouds simulated in this subsection use a cartesian coordinate system $(x,y,z)$. To generate the point clouds, $(\phi, z)$ are used as the grid space where $\phi_{i} = -\pi+\frac{2\pi i}{I}, z_j = \frac{j}{J}, i = 1,2,\ldots,I, j = 1,2,\ldots,J$. In this study, we set $I=50, J=50$. The equations for simulating the water drop point clouds are 
\begin{equation}
x(\phi, z)=\left(a(1+cos(b\phi))(1+sin(c\pi z))+d(-z^{2}+z)\right)cos(\phi)
\end{equation}
\begin{equation}
    y(\phi, z)=\left(a(1+cos(b\phi))(1+sin(c\pi z))+d(-z^{2}+z)\right)sin(\phi)
\end{equation}\\
where $\phi \in[-\pi, \pi], z \in[0,1]$. Each simulated point cloud can thus be stored as a matrix of shape ${I\times J\times 2}$. As can be seen in the simulation equations, there are 4 control variables that can control the shape of the simulated point cloud, and they are 1. $a$ controls the base radius of the water drop 2. $b$ controls the horizontal trigonometric radius pattern of the water drop 3. $c$ controls the vertical trigonometric radius pattern of the water drop 4. $d$ controls the magnitude of the extra quadratic curvature of the side surface. Each set of the 4 process variables corresponds to one shape of water drop point cloud. Assume the number of data collected is $N$, then the input tensor $X$ (process variables) would have shape $N\times 4$ and the output tensor $Y$ (point cloud coordinates) would have shape $N\times I\times J\times 2$. \\

\begin{figure}[htbp]
\includegraphics[scale=0.5]{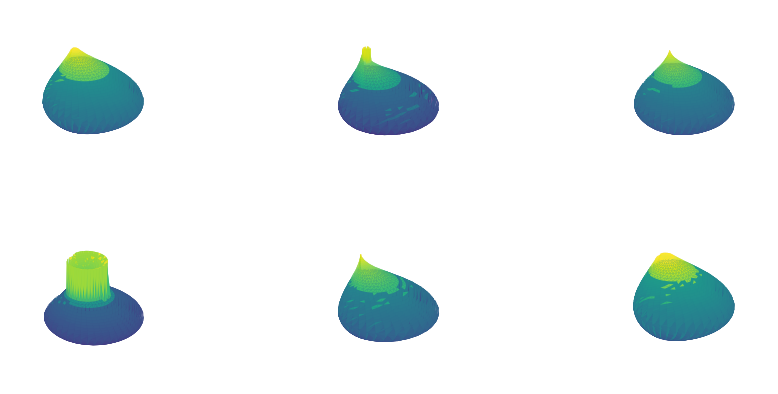}
\centering
\caption{\label{fig:simu1plot}Generated samples of the water drop point clouds}
\end{figure}

Since our algorithm is superior in dealing with non-linearity in the data, we simulate the control variables as $(a,b,c,d) = (U_1,0.5U_2,U_3,U_4)$ where $U_1,U_2,U_3,U_4\overset{i.i.d.}{\sim}$ Uniform$(1,2)$. The values and ranges are chosen so that the generated shapes of the simulated water drops are valid while non-linear and volatile enough to avoid any kind of linear approximation. After the simulation, i.i.d. Gaussian noises $N(0,\sigma^2)$ are added to the generated tensors. Figure \ref{fig:simu1plot} shows 6 randomly generated water drops following the above scheme.\\

\begin{figure}[htbp]
\includegraphics[scale=0.35]{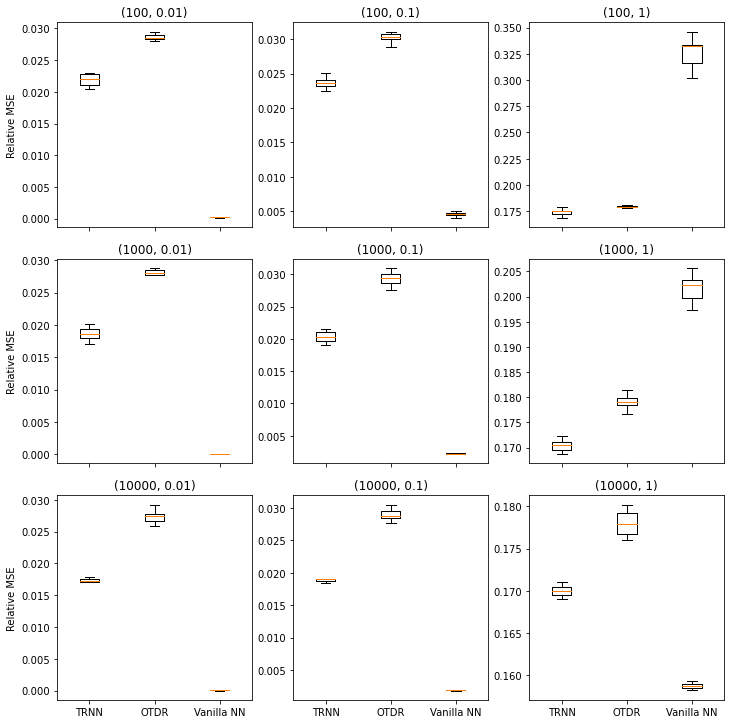}
\centering
\caption{\label{fig:simu1box} RMSE comparison with varying $(N,\sigma)$ for water drops}
\end{figure}

We compare our algorithm with the SOTA method OTDR and vanilla neural networks. To evaluate and compare the performances, we repeat 100 times, each time generated $N$ training data and 1000 test data, ran our algorithm and benchmarks, and calculated the Relative Mean Square Error (RMSE) over the test data. Given the predicted water drops $\hat{Y}\in\mathbb{R}^{N\times I\times J}$ and the true cones $Y\in\mathbb{R}^{N\times I\times J}$ from test data, the RMSE is evaluated as 
\begin{equation*}
    RMSE = \frac{\|\hat{Y}-Y\|_F^2}{\|Y\|_F^2}
\end{equation*}
\\
The RMSE metrics evaluated each run are recorded to produce box plots. The running times of the two algorithms are recorded as well. Furthermore, to evaluate the sensitivity of our algorithm with respect to number of training samples $N$ and the noise level $\sigma$, we repeat the above evaluation process for each combination of $N=100,1000,10000$ and $\sigma = 0.01,0.1,1$ respectively. The box plots for the comparison of RMSE are shown in Figure \ref{fig:simu1box} and the table for the comparison of running time is shown in Table \ref{table:simu1}.\\

\begin{table}[htbp]
\centering
\scalebox{0.9}{
\begin{tabular}{c|ccc|ccc|ccc}

\hline\(N\) & \multicolumn{3}{c}{ 100} & \multicolumn{3}{c}{1000} & \multicolumn{3}{c}{ 10000 } \\
\hline\(\delta\) & 0.01 & 0.1 & 1 & 0.01 & 0.1 & 1 & 0.01 & 0.1 & 1 \\
\hline TRNN & \(37.55\) & \(37.52\) & \(37.66\) & \(\textbf{94.36}\) & \(\textbf{94.74}\) & \(\textbf{94.70}\) & \(1130.48\) & \(1134.20\) & \(1134.04\) \\
OTDR & \(\textbf{19.64}\) & \(\textbf{19.52}\) & \(\textbf{19.50}\) & \(96.19\) & \(96.45\) & \(96.50\) & \(\textbf{949.54}\) & \(\textbf{951.35}\) & \(\textbf{955.27}\) \\
Vanilla NN & \(1157.35\) & \(1157.95\) & \(1171.74\) & \(1452.07\) & \(1460.21\) & \(1482.03\) & \(5045.17\) & \(5049.91\) & \(5107.43\) \\
\hline
\end{tabular}
}
\caption{Median running time (in sec) comparison with varying $(N,\sigma)$ for water drops}
\label{table:simu1}
\end{table}

As can be seen from the RMSE comparison, our algorithm always performs better than the tensor-based benchmark OTDR regardless of sample sizes or noise levels. This is as expected since we created extremely non-linear data in this simulation study. Comparing to the vanilla neural networks, we can notice that when the training samples are as small as 100-1000 and the noise level are as large as 1, our algorithm turns out to beat the vanilla neural networks. This is because the small sample size and large noises cause the vanilla neural networks to overfit. The drawback of the vanilla neural networks is its enormous number of parameters readily leading to overfitting, and we propose our tensor-based regression neural networks just to solve this high dimensionality issue. Under the manufacturing or many other traditional industrial application areas there is usually a lack of large amount of data and data often comes with large noises. This can cause the vanilla neural networks to overfit and our method demonstrates its competence. \\

In terms of the running time, when the number of samples is as small as 100, our method runs slower than OTDR. However, when sample sizes get larger, our running times become comparable to OTDR. Comparing to vanilla neural networks, due to our significantly smaller number of parameters, TRNN runs much faster than the its neural network counterpart, especially when the number of training samples are small where the RMSE results also show superiority.\\

It is interesting to notice that our TRNN can generalize well on the test data with small training data and running time. This is because our data and underlying model, though possessing strong non-linearity, are still not as complex as the most popular application areas including Computer Vision or Natural Language Processing. Since our task is not that hard, the training data not necessarily needs to be very large to generalize well. The same holds for small training time.\\

\subsection{Helicoid Point Cloud Simulation}
\label{sec:{Pointcloud}}

The simulation of the helicoid point clouds in this subsection involves mapping a 3D cylindrical coordinate system $(r,\phi,z)$ to a 3D cartesian coordinate system $(x,y,z)$. A helicoid has spiral structure such that for each height level $z\in[0,1]$, there is only one direction in $\phi$ that has points arranged in a half-line (arm) from the helicoid axis with arm locations $r\in [0,1]$. We choose this specific angle $\phi = \alpha z$ where $\alpha$ is a process parameter that controls the periodicity of the spiral structure with respect to the height $z$. To generate the helicoid, grid space $(r, z)$ are created over $r_{i} = \frac{i}{I}, z_j = \frac{j}{J}, i = 1,2,\ldots,I, j = 1,2,\ldots,J$ where $r$ represents the point location on the helicoid arm and $z$ represents the point height. For each pair of $(r,z)$, its cartesian coordinates $(x,y,z)$ are recorded as functional values at that pair. Since $z$ in $(x,y,z)$ is identical from $z$ in the grid space $(r,z)$, we omit it and collect all the $(x,y)$ coordinates over the grid to create a single functional response of shape $2\times I\times J$. The corresponding tensor response $Y\in \mathbb{R}^{N\times 2\times I\times J}$, where $N$ is the number of helicoid samples. In this study, we set $I=50, J=50$. We added extra non-linearity into the model by adding periodicity in the arm length varying with height, and the equations for simulating the helicoid point cloud are \\
\begin{equation}
    x(r,z)= (c_1+c_2\cos{\beta z})r\cos{\alpha z}
\end{equation}
\begin{equation}
    y(r,z)= (c_1+c_2\cos{\beta z})r\sin{\alpha z}
\end{equation}
where $r\in [0,1], z\in [0,1], 0<c_2<c_1$. \\

\begin{figure}[htbp]
\includegraphics[scale=0.5]{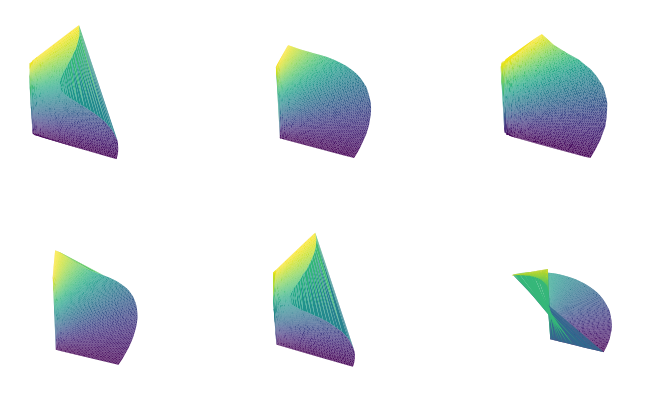}
\centering
\caption{\label{fig:simu2plot}Generated samples of the helicoid point clouds}
\end{figure}

 As can be seen in the simulation equations, there are 4 control variables that can control the shape of the helicoid point cloud, and they are 1. $c_1$ controls the base length of the helicoid arms 2. $c_2$ controls the varying range (amplitude) of the arm length 3. $\alpha$ controls the periodicity of arm angle 4. $\beta$ controls the periodicity of arm length. Each set of the 4 process variables corresponds to one shape of helicoid point cloud. To test on tensor-on-tensor regression, we further assume that instead of $\alpha,\beta$, we can only observe their corresponding profile predictors $x_3(z) = \cos{\alpha z}, x_4(z) = \cos{\beta z}$, together with two scalar predictors $x_1 = c_1$ and $x_2 = c_2$. To form the 2 scalars and 2 profiles into one input tensor $X$, each of the predictors is first formed into a matrix $X_1 \in \mathbb{R}^{N\times 1},X_2 \in \mathbb{R}^{N\times 1},X_3 \in \mathbb{R}^{N\times 50},X_4 \in \mathbb{R}^{N\times 50}$ and we replicate the column of $X_1, X_2$ 50 times to produce matrices of shape $N\times 50$ before merging them together with $X_3,X_4$ to obtain $X\in \mathbb{R}^{N\times 4\times 50}$. This is the input for our algorithm TRNN. For the benchmark MTOT \citep{gahrooei2021multiple}, $X_1,X_2$ are together formed into $X_1\in \mathbb{R}^{N\times 2}$ and $X_3,X_4$ are together formed into $X_2\in \mathbb{R}^{N\times 2\times 50}$ as inputs. We simulate the control variables as $(c_1, c_2, \alpha, \beta) = (5U_1,2U_2,3U_3,4U_4)$ where $U_1,U_2,U_3,U_4\overset{\text{i.i.d.}}{\sim} $ Uniform$(0.5,1.5)$. After the simulation, i.i.d. Gaussian noises $N(0,\sigma^2)$ are added to the generated tensors. Figure \ref{fig:simu2plot} shows 6 randomly generated helicoids following the above scheme.\\

\begin{figure}[htbp]
\begin{minipage}[t]{0.5\linewidth}
\centering
\includegraphics[scale=0.3]{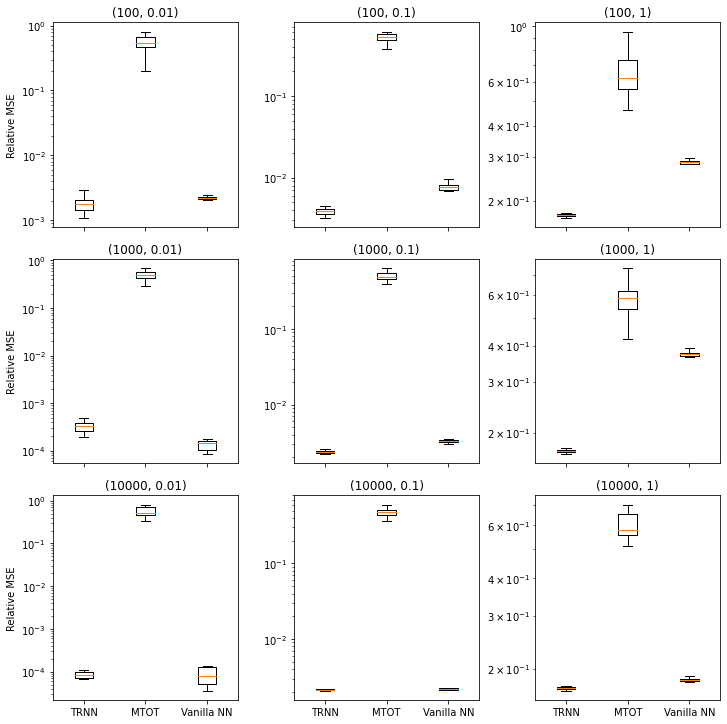}
\end{minipage}%
\begin{minipage}[t]{0.5\linewidth}
\centering
\includegraphics[scale=0.3]{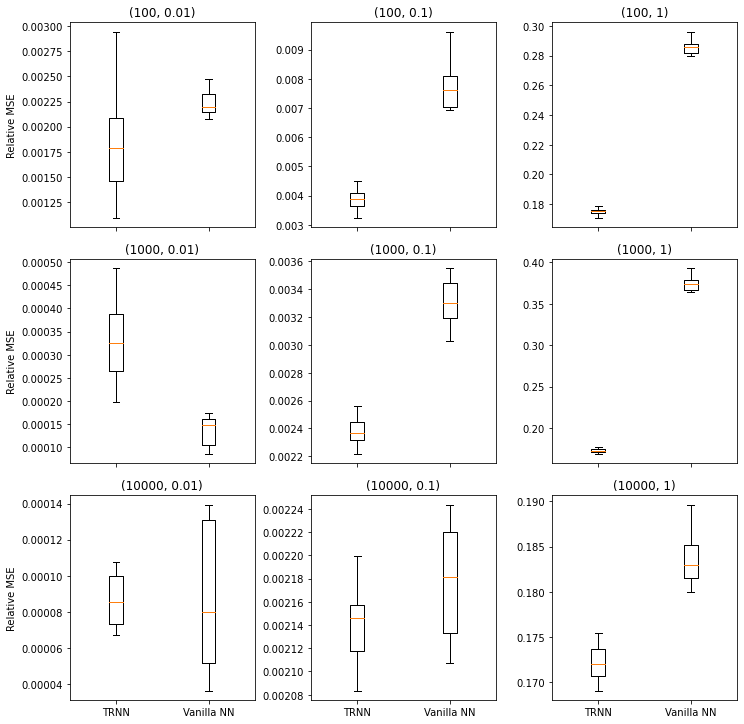}
\end{minipage}%
\caption{\label{fig:simu2box}RMSE comparison with varying $(N,\sigma)$ for helicoids (Left: log scale, Right: original scale, neural methods only)}
\end{figure}

We compare our algorithm with the SOTA tensor-based method MTOT and vanilla neural networks. To evaluate and compare the performances, we repeat 100 times, each time generated $N$ training data and 1000 test data, ran our algorithm and benchmarks, and calculated the Relative Mean Square Error (RMSE) over the test data. The RMSE metrics evaluated each run are recorded to produce the box plots. The running times of the two algorithms are recorded as well. Furthermore, to evaluate the sensitivity of our algorithm with respect to number of training samples $N$ and the noise level $\sigma$, we repeat the above evaluation process for each combination of $N=100,1000,10000$ and $\sigma = 0.01,0.1,1$ respectively. The box plots for the comparison of RMSE are shown in Figure \ref{fig:simu2box} and the table for the comparison of running time is shown in Table \ref{table:simu2}.\\

\begin{table}[htbp]
\centering
\scalebox{0.9}{
\begin{tabular}{c|ccc|ccc|ccc}

\hline\(N\) & \multicolumn{3}{c}{ 100} & \multicolumn{3}{c}{1000} & \multicolumn{3}{c}{ 10000 } \\
\hline\(\delta\) & 0.01 & 0.1 & 1 & 0.01 & 0.1 & 1 & 0.01 & 0.1 & 1 \\
\hline TRNN & \(\textbf{54.76}\) & \(\textbf{44.55}\) & \(\textbf{44.33}\) & \(\textbf{86.04}\) & \(\textbf{78.83}\) & \(\textbf{78.31}\) & \(\textbf{1866.92}\) & \(\textbf{1542.97}\) & \(\textbf{1222.17}\) \\
MTOT & \(338.55\) & \(290.12\) & \(286.01\) & \(1249.55\) & \(1123.04\) & \(1103.98\) & \(14083.38\) & \(11469.47\) & \(11967.72\) \\
Vanilla NN & \(1422.46\) & \(1373.51\) & \(1177.60\) & \(1502.79\) & \(1500.69\) & \(1509.49\) & \(5076.98\) & \(5019.80\) & \(5071.23\) \\
\hline
\end{tabular}
}
\caption{Median running time (in sec) comparison with varying $(N,\sigma)$ for helicoids}
\label{table:simu2}
\end{table}

As can be seen from the RMSE comparison, in all cases, even when the number of training samples is as small as 100, our algorithm turns out to be largely superior to the benchmark MTOT. This is because the model used to generate the helicoids in this study is extremely non-linear and involves many multiplications of nonlinear functions, so purely linear methods like MTOT fail to perform well. It is also interesting to note that the RMSE of MTOT is more consistent over different hyperparameters $N$ and $\sigma$ than our method, kept at some high level. This is because the main bottleneck of MTOT is the linear assumption of the underlying relationship. Having more data or smaller noise level does not help the training. For our method, with more data and less noise, we can see the performance grows much better. When compared to vanilla neural networks, except for when sample size is large and noise level is small, our method is always much better. This can be expected as vanilla neural networks have way large number of parameters and tend to have overfitting at the presence of high noise or small sample size. In the real world applications, especially in the manufacturing setting, data size is usually small with large noises, and this is when our method gains much competence.\\ 

The comparison of the running time shows that under all the circumstances our method is much faster than the benchmarks. 

\section{Case Studies}
\label{sec:Casestudy}

In this section, we did two case studies to evaluate the performance of our algorithm. The first study focused on the scalar-on-tensor regression problem occurred in the point cloud modeling in a turning process, and we compared our approach with the corresponding SOTA tensor-based method OTDR. The second focused on the tensor-on-tensor regression problem arose in the curve-on-curve modeling in an emission control system, and we compared our approach with the corresponding SOTA tensor-based method MTOT. Curve-on-curve regression is another popular application area in the industry. Also, to have a fair comparison, we compare our performance with the benchmark from neural networks side, using the simplest vanilla neural networks (Vanilla NN).

\subsection{Titanium Alloy Cylinder}
\label{sec:Titanium}

Ti-6Al-4V is a type of titanium alloy that is ideal for mechanical components in aerospace due to its high specific strength, corrosion and erosion resistance, high fatigue strength \citep{peters2003titanium}. However, titanium alloys are the type of material that is difficult to machine in general \citep{pramanik2014problems,pervaiz2014influence}. Various research has been conducted in order to investigate the best cutting parameters (like cutting speed and cutting depth) to produce the high quality titanium alloy parts \citep{chauhan2012optimization,khanna2015design}. To manufacture parts with shapes close to the nominal, models can be built between the process parameters and machined product shapes, and subsequently be used to find the best cutting setting. In this way, product quality can also be controlled by monitoring the parameters. \\

In this section, surface roughness data of Ti-6Al-4V cylinders machined from a lathe turning process is used to evaluate our method. This dataset is from the experiment conducted in \cite{pacella2018multilinear}. During the lathe turning process, 20-mm-diameter raw bars were machined to bars of 16.8 mm diameter by two cutting steps. The second cutting step is called finishing operation, and a $3^2$ full factorial experiment is conducted during this operation. The operation involves two process variables and each of them has 3 levels: cutting speed has levels 65, 70, and 80 m/min, and cutting depth has levels 0.4, 0.8, and 1.2 mm. For each combination of levels, the corresponding experiment was replicated 10 times, and the control variables are recorded in a input matrix $X\in \mathbb{R}^{90\times 2}$. To measure the response cylindrical surfaces, a Coordinate Measuring Machine (CMM) \citep{dowllng1997statistical} with a touch trigger probe was used. For each alloy sample, the measurements were taken on a cylindrical grid of size $210\times64$, on the 210 equally spaced cross-sections along the 42 mm bar and 64 equally spaced angular positions along each section. The probe approached the manufactured cylinder on each grid point, and recorded the coordinates once touching the surface. After each surface was measured, its substitute geometry (a reference nominal cylinder) was computed from the ideal cylinder features by least square, and the radial derivations of the measured cylinder from the reference one on the grid were used as the final measurements $Y\in \mathbb{R}^{90\times 210\times64}$. Registration was also conducted on the data to deal with form tolerances \citep{colosimo2011analyzing}. Figure \ref{fig:case1plot} gives examples of the average response surfaces under different cutting conditions. To better visualize the differences, the radial deviations are scaled 250 times larger. It is clear that the cylindrical shapes are affected by both of the process variables. In particular, the decreases of both the cutting speed and depth can lead to larger taper axial form error \citep{henke1999methods}.\\

\begin{figure}[htbp]
\centering
\begin{minipage}{0.3\textwidth}
    \includegraphics[scale=0.5]{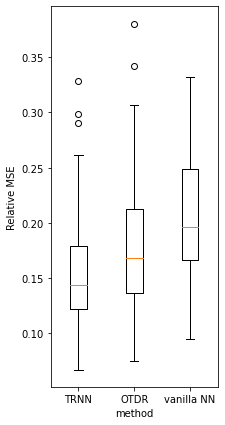}
    \centering
   \end{minipage}\hspace{0.2cm}
   \begin{minipage}{0.3\textwidth}
    \includegraphics[scale=0.5]{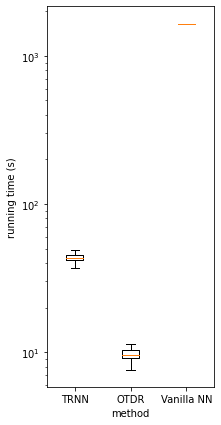}
    \centering
   \end{minipage}
   \caption{\label{fig:case1box} Comparison of RMSE and running time for titanium alloy cylinders}
\end{figure}

We compare our algorithm with the SOTA tensor-based method OTDR and vanilla neural networks. To evaluate and compare the performance, we repeat 100 times. Each time we randomly sample 85 out of 90 as training data and the remaining 5 as the test data, do the training, record the running time and calculate the RMSE on the test data. The box plots for the comparison of RMSE and running time are shown in Figure \ref{fig:case1box}. As can be seen from the RMSE comparison, our algorithm turns out to be superior to the benchmark OTDR even in the case where the number of collected samples is as small as 90. Because of the small training size the improvement is not very significant, but we still achieved a $15\%$ improvement of the median RMSE from 0.1684 to 0.1435, and a $11\%$ decrease of the standard deviation from 0.0571 to 0.0506. In terms of vanilla neural networks, we achieved a $27\%$ improvement of the median RMSE from 0.1959 to 0.1435, and a $23\%$ decrease of the standard deviation from 0.0661 to 0.0506. From simulation 1 we know that in case of small training size or large noise, our method will outperform vanilla neural networks, and this case also confirms it. \\

In the meanwhile, the training of our method took longer than the OTDR, 43.37 seconds comparing to the 9.55 seconds in median. This is again because of the small training size. However, the training time is still acceptable, especially in our offline setting. The running time of vanilla neural networks is even large, 1638.12s in median. 

\subsection{Vehicle Engine Sensors Related to Air/Fuel Ratio}
\label{sec:Engine}

In this section, HD multichannel profile data related to the $\lambda$-undershoot fault collected from a diesel vehicle engine is used to evaluate our method. The dataset is from the case study in \cite{pacella2018unsupervised}.\\

The NOx Storage Catalyst (NSC) is a type of emission control systems that is capable of reducing the NOx in the exhaust gas from vehicles. The exhaust gas is processed in two stages. During the adsorption stage, the NOx molecules are captured by an adsorber with built-in zeolites coated catalytic converter; After the adsorber becomes saturated, the adsorbed NOx is decomposed by the catalyst within 30 to 90 seconds, and this is called regeneration stage. The second phase requires the maintenance of a rich air-to-fuel environment during the combustion, and an engine control unit (ECU) is programmed to maintain this. The maintenance is facilitated by measurements of relative air/fuel ratio ($\lambda$) from sensors in the upstream NSC. As the indicator of a correct regeneration phase, the $\lambda$ signal 
is in most of the time held within the range [0.92,0.95]. However, when faults occur, the $\lambda$ value would fall below an acceptable threshold, usually jump to [0.8, 0.9]. This type of fault is called $\lambda$-undershoot, which can diminish the performance of NSC and should thus be prevented \citep{pacella2018unsupervised}. Though the $\lambda$ value can be easily monitored, the reason behind the emergence of the undershoot and how to adjust the true faulty engine operation to correct it is hard to identify. This calls for the development of models that can predict the $\lambda$ signal given other related engine signals to support the diagnosis and the fix of the fault. \\

\begin{figure}[htbp]
\includegraphics[scale=0.45]{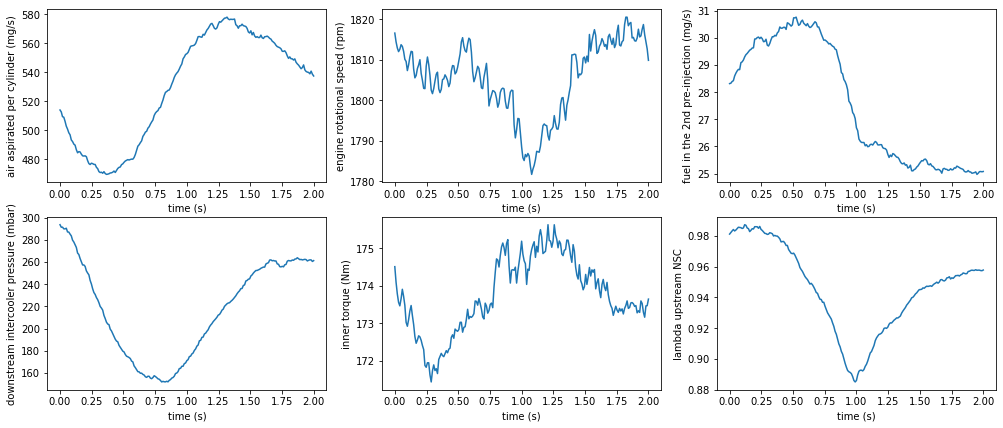}
\centering
\caption{\label{fig:case2plot} Median curves of the input and output signals related to $\lambda$}
\end{figure}

Numerous on-broad engine sensors are installed in modern vehicles, generating HD real-time profiles reflecting the vehicle status. The signals related to the undershoot issue include engine inner torque, rotational speed, and quantity of fuel injected \citep{gahrooei2021multiple}. Our dataset contains 342 faulty samples and for each a $\lambda$-undershoot event has occurred. Each entry consists of one response signal $\lambda$ and 5 input signals from the regeneration stage known to be closely related to $\lambda$. The signals are illustrated in Figure \ref{fig:case2plot}. These signals are all measured over a 2s time window, which is selected upon the triggering of a $\lambda$-undershoot event on the $\lambda$ signal. Specifically, the midpoint of the time window is set to be the moment the undershoot happened. With a sampling frequency of around 100Hz, 203 measurements were made within the 2s time window. \\

\begin{figure}[htbp]
\centering
\begin{minipage}{0.3\textwidth}
    \includegraphics[scale=0.5]{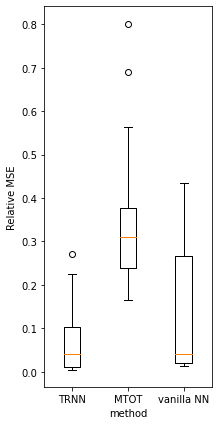}
    \centering
   \end{minipage}\hspace{0.2cm}
   \begin{minipage}{0.3\textwidth}
    \includegraphics[scale=0.5]{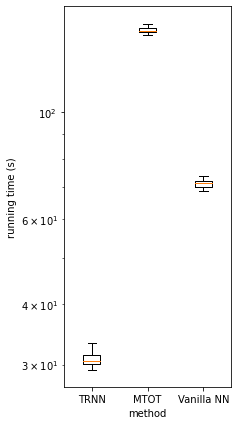}
    \centering
   \end{minipage}
   \caption{\label{fig:case2box} Comparison of RMSE and running time for $\lambda$ undershoot problem}
\end{figure}

To compare our approach with the SOTA method MTOT, we create input tensor $X\in \mathbb{R}^{342\times 5\times203}$ and output tensor $Y\in \mathbb{R}^{342\times203}$. We repeat the evaluation procedure 100 times, each time randomly sampling 62 out of 342 as test data and the remaining 280 as training data (to achieve a train-test-split of about 80$\%$), do the training, record the running time and calculate the RMSE on the test data. The box plots for the comparison of RMSE and running time are shown in Figure \ref{fig:case2box}. \\

As can be seen from the RMSE comparison, our algorithm turns out to perform much better than MTOT, boosting the performance almost 10 times from 0.3110 to 0.03985 in terms of the median RMSE. The standard deviation also observed an over $50\%$ decrease from 0.1453 to 0.06735. This enormous increase in performance indicates the strong non-linearity in the relationship between the $\lambda$ and input signals, which failed to be captured by the linear model. At the same time, the running time of our method is much faster than the MTOT, 30.56 seconds comparing to the 146.96 seconds in median. This is because our method uses gradient descent as a cheap optimization method to approximate the global optimum. The MTOT method, on the other hand, involves lots of expensive tensor operations and matrix inversions.\\

Our method outperforms vanilla neural networks in RMSE, improving the median from 0.04053 to 0.03985. Though the improvement in median is not very significant, if we focus on the box plot instead of sore median, we can see the RMSE of vanilla neural networks are overall higher than our method in terms of each experiment replication. This is reflected in the standard deviation of RMSE, where our method is 0.06735 comparing to 0.1533 of vanilla neural networks. The running time of our method also more than halved that of vanilla neural networks (71.34 seconds).

\section{Conclusion}
\label{sec:Conclusion}

HD data, including profiles, images and point clouds, prevail in many areas such as manufacturing and mechanical engineering as a result of the advancement of modern sensing and metrology technology. Constructing regression models between HD data is an important research topic for the understanding of system behavior and subsequent diagnosis. Tensor techniques are widely used in the literature for process modeling involving HD data represented as tensors. Nevertheless, current methodologies fail to capture the complex nonlinear associations between HD data in the real-world processes. To overcome it, we proposed a novel tensor-on-tensor regression neural networks framework for the modeling of these nonlinear relations. The simulation studies indicated that our method in general outperforms the SOTA tensor and neural networks methodologies. Eventually, two real case studies, one on the point cloud modeling in a turning process and another on the curve-on-curve regression in an emission control system, were established to evaluate our proposed method. The results further demonstrated the superiority of our approach in practical performances.\\

Several potential research directions can be pursued by future researchers. First, the idea of our algorithm can be adapted to other industrial applications such as process monitoring and optimization. Second, our nonlinear method treating HD data as tensors still fails to consider the type of data which cannot be formed into tensors, such as unstructured point clouds. The investigation of the nonlinear modeling involving unstructured point cloud can be another challenging but worthwhile topic.

\bigskip




\bibliographystyle{chicago}
\bibliography{main}

\end{document}